# Learning to Predict from Textual Data

**Kira Radinsky**                                                KIRAR@CS.TECHNION.AC.IL
**Sagie Davidovich**                                            MESAGIE@GMAIL.COM
**Shaul Markovitch**                                         SHAULM@CS.TECHNION.AC.IL
*Computer Science Department*
*Technion—Israel Institute of Technology*
*Haifa 32000, Israel*

## Abstract

Given a current news event, we tackle the problem of generating plausible predictions of future events it might cause. We present a new methodology for modeling and predicting such future news events using machine learning and data mining techniques. Our Pundit algorithm generalizes examples of causality pairs to infer a causality predictor. To obtain precisely labeled causality examples, we mine 150 years of news articles and apply semantic natural language modeling techniques to headlines containing certain predefined causality patterns. For generalization, the model uses a vast number of world knowledge ontologies. Empirical evaluation on real news articles shows that our Pundit algorithm performs as well as non-expert humans.

## 1. Introduction

Causality has been studied since antiquity, e.g., by Aristotle, but modern perceptions of causality have been most influenced, perhaps, by the work of David Hume (1711–1776), who referred to causation as the strongest and most important associative relation, that which lies at the heart of our perception and reasoning about the world, as "it is constantly supposed that there is a connection between the present fact and that which is inferred from it."

Causation is also important for designing computerized agents. When an agent, situated in a complex environment, plans its actions, it reasons about future changes to the environment. Some of these changes are a result of its own actions, but many others are a result of various chains of events not necessarily related to the agent. The process of observing an event, and reasoning about future events that might be *caused* by it, is called *causal reasoning*.

In the past, computerized agents could not operate in complex environments due to their limited perceptive capabilities. The proliferation of the World Wide Web, however, changed all that. An intelligent agent can act in the virtual world of the Web, perceiving the current state of the world through extensive sources of textual information, including Web pages, tweets, news reports, and online encyclopedias, and performing various tasks such as searching, organizing, and generating information. To act intelligently in such a complex virtual environment, the agent must be able to perceive the current state and reason about future states through causal reasoning. Such reasoning ability can be extremely helpful in conducting complex tasks such as identifying political unrest, detecting and tracking social trends, and generally supporting decision making by politicians, businesspeople, and individual users.

While many works have been devoted to extracting information from text (e.g., Banko, Cafarella, Soderl, Broadhead, & Etzioni, 2007; Carlson, Betteridge, Kisiel, Settles, Hruschka, & Mitchell, 2010), little has been done in the area of causality extraction, with the works of Khoo, Chan, and Niu (2000) and Girju and Moldovan (2002) being notable exceptions. Furthermore, the algorithms developed for causality extraction try to *detect* causality and cannot be used to *predict* it, that is, to generate new events the given event might cause.

Our goal in this paper is to provide algorithms that perform causal reasoning, in particular causality prediction, in textually represented environments. We have developed a causality learning and prediction algorithm, Pundit, that, given an event represented in natural language, predicts future events it can cause. Our algorithm is trained on examples of causality relations. It then uses large ontologies to generalize over the causality pairs and generate a prediction model. The model is represented by an abstraction tree, that, given an input cause event, finds its most appropriate generalization, and uses learned rules to output predicted effect events.

We have implemented our algorithm and applied it to a large collection of news reports from the last 150 years. To extract training examples from the news corpus, we do not use correlation, by means of which causality is often misidentified. Instead, we use *textual causality patterns* (such as "X because Y" or "X causes Y"), applied to news headlines, to identify pairs of structured events that are supposedly related by causality. The result is a semantically-structured causality graph of 300 million fact nodes connected by more than one billion edges. To evaluate our method, we tested it on a news archive from 2010, which was not used during training. The results are judged by human evaluators.

To give some intuition about the type of predictions the algorithm generates, we present here two examples of actual predictions made by our system. First, given the event "Magnitude 6.5 earthquake rocks the Solomon Islands," the algorithm predicted that "a tsunami-warning will be issued for the Pacific Ocean." It learned this from past examples on which it was trained, one of which was
⟨*7.6 earthquake strikes island near India, tsunami warning issued for Indian Ocean*⟩.
Pundit was able to infer that an earthquake occurring near an island would result in a tsunami warning being issued for its ocean. Second, given the event "Cocaine found at Kennedy Space Center," the algorithm predicted that "a few people will be arrested." This was partially based on the past example ⟨*police found cocaine in lab → 2 people arrested*⟩.

The contributions of this work are threefold: First, we present novel and scalable algorithms for generalizing causality pairs to causality rules. Second, we provide a new method for using casualty rules to predict new events. Finally, we implement the algorithms in a large scale system and perform an empirical study on realistic problems judged by human raters. We make the extracted causality information publicly available for further research in the field [1].

---

1. http://www.technion.ac.il/~kirar/Datasets.html

## 2. Learning and Predicting Causality

In this section, we describe the Pundit algorithm for learning and predicting causality. We start with an overview of the learning and prediction process. During training, the learning algorithm receives causality event pairs, extracted from historical news archives (Section 3). The algorithm then generalizes over the given examples using world knowledge and produces an abstraction tree (AT)(Section 2.4). For each node in the AT, a prediction rule is generated from the examples in the node (Section 2.5). Then, during the prediction phase, the algorithm matches the given new event to nodes in the AT, and the associated rule is applied on it to produce possible effect events (Section 2.6). Those events are then filtered (Section 2.7) and an effect event output. The output event itself is also given in natural language, in sentence-like form. The process is illustrated in Figure 1.

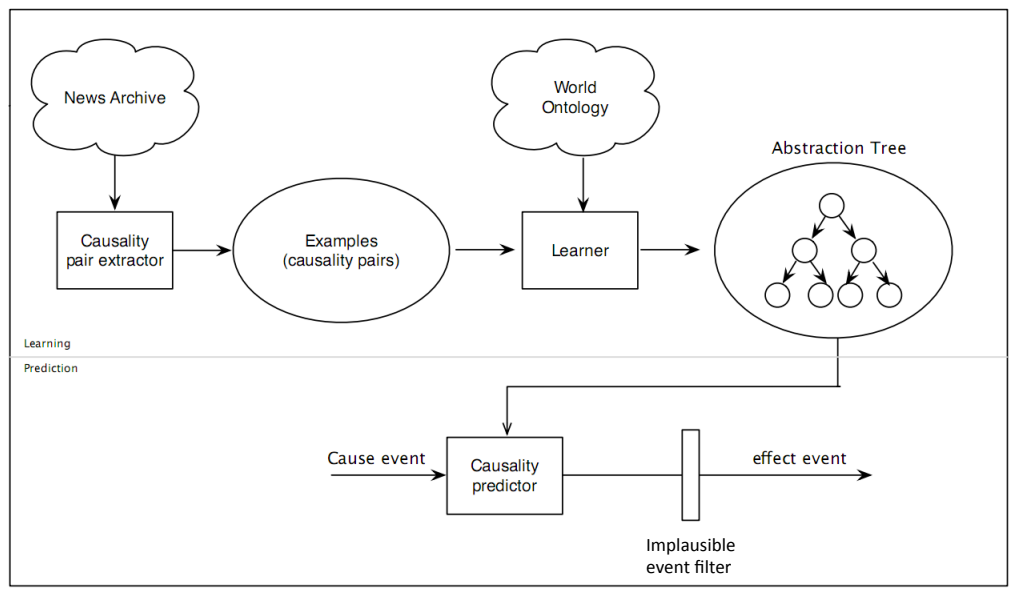

Figure 1: Structure of the Pundit prediction algorithm

### 2.1 Event Representation

The basic element of causal reasoning is an event. The Topic Tracking and Detection (TDT) community (Allan, 2002) has defined an event as "a particular thing that happens at a specific time and place, along with all necessary preconditions and unavoidable consequences."

Other philosophical theories consider events as exemplifications of properties by objects at times (Kim, 1993). For example, Caesar's death at 44 BC is Caesar's exemplification of the property of dying at time 44 BC. Those theories impose structure on events, where a change in one of the elements yields a different event. For example, Shakespear's death is a different event from Caesar's death, as the objects exemplifying the property are different. In this section, we will discuss a way to represent events following Kim's (1993) exemplification theory that will allow us to easily compare them, generalize them, and reason about them.

There are three common approaches for textual event representation: The first approach describes an event at sentence level by running text or individual terms (Blanco, Castell, & Moldovan, 2008; Sil, Huang, & Yates, 2010). Event similarity is treated as a distance metric between the two events' bag of words. While such approaches can be useful, they often fail to perform fine-grained reasoning. Consider, for example, three events: "US Army bombs a warehouse in Iraq," "Iraq attacks US base," and "Terrorist base was attacked by the US Marines in Kabul." Representing these events by individual terms alone might yield that the first two are more similar than the first and the last as they have more words in common. However, such approaches disregard the fact the actors of the first and last event are military groups and that Kabul and Iraq are the event locations. When these facts are taken into account, the first and last events are clearly more similar than the first and second.

The second approach describes events in a syntax-driven manner, where the event text is transformed into syntax-based components, such as noun phrases (Garcia, 1997; Khoo et al., 2000; Girju & Moldovan, 2002; Chan & Lam, 2005). In our example, this representation again erroneously finds the second and third events to be most similar due to the syntactic similarity between them. Using the first two approaches, it is hard to make practical generalizations about events or to compare them in a way that takes into account all the semantic elements that compose them.

The third approach is semantic (similar to the representation in Cyc; Lenat & Guha, 1990), and maps the atomic elements of each event to semantic concepts. This approach provides grounds for canonic representation of events that are both comparable and generalizable. In this work, we follow the third approach and represent events semantically.

Given a set of entities $O$ that represent physical objects and abstract concepts in the real world (e.g., people, instances, and types), and a set of actions $P$, we define an event as an ordered set $e = \langle P, O_1, \ldots, O_4, t \rangle$, where:

1. $P$ is a temporal action or state that the event's objects exhibit.

2. $O_1$ is the *actor* that performed the action.

3. $O_2$ is the *object* on which the action was performed.

4. $O_3$ is the *instrument* with which the action was performed.

5. $O_4$ is the *location* of the event.

6. $t$ is a time-stamp.

For example, the event "The U.S Army destroyed a warehouse in Iraq with explosives," which occurred on October 2004, is modeled as: Destroy (Action); U.S Army (Actor); warehouse (Object); explosives (Instrument); Iraq (Location); October 2004 (Time). This approach is inspired by Kim's (1993) property-exemplification of events theory.

## 2.2 Learning Problem Definition

We treat causality inference as a supervised learning problem. Let $Ev$ be the universe of all possible events. Let $f : Ev \times Ev \to \{0, 1\}$ be the function

$$f(e_1, e_2) = \begin{cases} 1 & \text{if } e_1 \text{ causes } e_2, \\ 0 & \text{if otherwise.} \end{cases}$$

We denote $f^+ = \{(e_1, e_2) | f(e_1, e_2) = 1\}$. We assume we are given a set of possible positive examples $E \subseteq f^+$.

Our goal is not merely to *test* whether a pair of events is a plausible cause-effect pair by $f$, but to *generate* for a given event $e$ the events it can cause. For this purpose we define $g : Ev \to 2^{Ev}$ to be $g(e) = \{e' | f(e, e') = 1\}$; that is, given an event, output the set of events it can cause. We wish to build this predictor $g$ using the examples $E$.

Learning $f$ from $E$ could have been solved by standard techniques for concept learning from positive examples. The requirement to learn $g$, however, presents the challenging task of structured prediction from positive examples.

## 2.3 Generalizing Over Objects and Actions

Our goal is to develop a learning algorithm that automatically produces a causality function based on examples of causality pairs. The inferred causality function should be able to predict the outcome of a given event, even if it was never observed before. For example, given the training examples ⟨earthquake in Turkey, destruction⟩ and ⟨earthquake in Australia, destruction⟩, and a current new event of "earthquake in Japan," a reasonable prediction would be "destruction." To be able to handle such predictions, we must endow our learning algorithm with generalization capacity. For example, in the above scenario, the algorithm must be able to generalize Australia and Turkey to countries, and to infer that earthquakes in countries might cause destruction. This type of inference and the knowledge that Japan is also a country enables the algorithm to predict the effects of new events using patterns in the past.

To generalize over a set of examples, each consisting of a pair of events, we perform generalization over the components of these events. There are two types of components – objects and actions.

To generalize over objects, we assume the availability of a semantic network $G_o = (V, E)$, where the nodes $V \subseteq O$ are the objects in our universe, and the labels on the edges are relations such as isA, partOf and CapitalOf. In this work, we consider one of the largest semantic networks available, the LinkedData ontology (Bizer, Heath, & Berners-Lee, 2009), which we describe in detail in Section 3.

We define two objects to be similar if they relate to a third object in the same way. This relation can be a label or a sequence of labels in the semantic network. For example,

Paris and London will be considered similar because their nodes are connected by the path $\xrightarrow{Capital-of} \xrightarrow{In-continent}$ to the node Europe. We now formally define this idea.

**Definition 1.** *Let $a, b \in V$. A sequence of labels $L = l_1, \ldots, l_k$ is a **generalization path** of $a, b$, denoted by GenPath(a,b), if there exist two paths in $G$, $(a, v_1, l_1), \ldots (v_k, v_{k+1}, l_k)$ and $(b, w_1, l_1), \ldots (w_k, w_{k+1}, l_k)$, s.t. $v_{k+1} = w_{k+1}$.*

Overgeneralization of events should be avoided – e.g., given two similar events, one occurring in Paris and one in London, we wish to produce the generalization "city in Europe" ($\xrightarrow{Capital-of} \xrightarrow{In-continent} Europe$) rather than the more abstract generalization "city on a continent" ($\xrightarrow{Capital-of} \xrightarrow{In-continent} \xrightarrow{IsA} Continent$). We wish our generalization to be as specific as possible. We call this *minimal generalization* of objects.

**Definition 2.** *The **minimal generalization path**, denoted by MGenPath(a, b), is defined as the set containing the shortest generalization paths. We denote $dist_{Gen}(a, b)$ as the length of the MGenPath(a, b).*

Path-based semantic distances such as the one above were shown to be successful in many NLP applications. For example, the semantic relatedness of two words was measured by means of a function that measured the distance between words in a taxonomy (Rada, Mili, Bicknell, & Blettner, 1989; Strube & Ponzetto, 2006). We build on this metric and expand it to handle events that are structured and can contain several objects from different ontologies.

To efficiently produce $MGenPath$, we designed an algorithm (described in Figure 2), based on dynamic programming, that computes the $MGenPath$ for all object pairs in $G$. For simplicity, we describe an algorithm that computes a single path for each two nodes $a$ and $b$, rather than the set of all shortest paths. At step 1 a queue that holds all nodes with the same generalization is initialized. At step 2, the algorithm identifies all nodes $(a, b)$ that have a common node $(c)$ connecting to them via the same type of edge $(l)$. $c$ can be thought of as a generalization of $a$ and $b$. The $Mgen$ structure maps a pair of nodes to their generalization ($Mgen.Gen$) and their generalization path ($MGen.Pred$). At step 3, in a dynamic programming manner, the algorithm iterates over all nodes $(a, b)$ in $Mgen$ for which we found a minimal generalization in previous iterations, and finds two nodes – one $(x)$ connecting to $a$ and one $(y)$ connecting to $b$ via the same type of edge $l$ (step 3.4). Thus, the minimal generalization of $x$ and $y$ is the minimal generalization of $a$ and $b$, and the path is the $MGenPath$ of $a, b$ with the addition of the edge type $l$. This update is performed in steps 3.4.1–3.4.4. Eventually, when no more nodes with minimal generalization can be expanded (i.e., the algorithm cannot find two nodes that connect to them via the same edge type), it stops and returns $Mgen$ (step 4). During prediction, if several $Mgen$ exists, we consider both during the prediction with their corresponding $MGenPath$.

We define a distance between actions using an ontology $G_p$, similarly to the way we defined distance between objects. Specifically, we use the VerbNet (Kipper, 2006) ontology, which is one of the largest English verb lexicons. It has mapping to many other online resources, such as Wordnet (Miller, 1995). The ontology is hierarchical and is based on a classification of the verbs to the Levin classes (Dang, Palmer, & Rosenzweig, 1998). This resource has been widely used in many natural language processing applications (Shi &

---

**Procedure** MINIMAL GENERALIZATION PATH$(G)$

(1) $Q \leftarrow$ new Queue

(2) **Foreach** $\{(a, c, l), (b, c, l) \in E(G) |$

$\qquad a, b, c \in V(AT), l \in Lables\}$

(2.1) $Mgen(a, b).Gen = c$

(2.2) $Mgen(a, b).Pred = l$

(2.3) $Mgen(a, b).Expanded = false$

(2.4) $Q.enqueue((a, b))$

(3) **While** $Q \neq \emptyset$:

(3.1) $(a, b) \leftarrow Q.dequeue()$

(3.2) **If** $Mgen(a, b).Expanded \neq true$:

$\quad Mgen(a, b).Expanded = true$

(3.4) **Foreach** $\{(x, a, l), (y, b, l) \in E(AT) |$

$\qquad x, y \in V(AT), l \in Lables\}$

(3.4.1) $Mgen(x, y).Gen = Mgen(a, b).Gen$

(3.4.2) $Mgen(x, y).Pred = Mgen(a, b).Pred || l$

(3.4.3) $Mgen(x, y).Expanded = false$

(3.4.4) $Q.enqueue((x, y))$

(4)**Return** $Mgen$

---

Figure 2: Procedure for calculating the minimal generalization path for all object pairs

Mihalcea, 2005; Giuglea & Moschitti, 2006). Using this ontology we describe the connections between verbs. Figure 10 shows a node in this ontology that generalizes the actions "hit" and "kick."

## 2.4 Generalizing Events

In order to provide strong support for generalization, we wish to find similar events that can be generalized to a single abstract event. In our example, we wish to infer that both ⟨earthquake in Turkey, destruction⟩ and ⟨earthquake in Australia, destruction⟩ are examples of the same group of events. Therefore, we wish to cluster the events in such a way that events with similar causes and effects will be clustered together. As in all clustering methods, a distance measure between the objects should be defined. Let $e_i = \langle P^i, O_1^i, \ldots, O_4^i, t^i \rangle$ and $e_j = \langle P^j, O_1^j, \ldots, O_4^j, t^j \rangle$ be two events. In the previous subsection we defined a distance function between objects (and between actions). Here, we define the similarity of two events $e_i$ and $e_j$ to be a function of distances between their objects and actions:

$$SIM(e_i, e_j) = f\big(dist_{Gen}^{G_p}(P^i, P^j), dist_{Gen}^{G_o}(O_1^i, O_1^j), \ldots, dist_{Gen}^{G_o}(O_4^i, O_4^j)\big), \qquad (1)$$

where, $dist_{Gen}^G$ is the distance function $dist_{Gen}$ in the graph $G$, and $f$ is an aggregation function. In this work, we mainly use the average as the aggregation function, but also analyze several alternatives.

Likewise, a similarity between two pairs of cause-effect events $\langle c_i, e_i \rangle$ and $\langle c_j, e_j \rangle$ is defined as:

$$SIM(\langle c_i, e_i \rangle, \langle c_j, e_j \rangle) = f\big(SIM(c_i, c_j), SIM(e_i, e_j)\big). \tag{2}$$

Using the similarity measure suggested above, the clustering process can be thought of as a grouping of the training examples in such a way that there is a low variance in their effects and a high similarity in their cause. This is similar to information gain methods where examples are clustered by their class. We use the HAC hierarchical clustering algorithm (Eisen, Spellman, Brown, & Botstein, 1998) as our clustering method. The algorithm starts by joining the closest event pairs together into a cluster. It then keeps repeating the process by joining the closest two clusters together until all elements are linked into a hierarchical graph of events we call an abstraction tree (AT). Distance between clusters is measured by the distance of their representative events. To allow this, we assign to each node in the AT a representative cause event, which is the event closest to the centroid of the node's cause events. During the prediction phase, the input cause event will be matched to one of the created clusters, i.e., closest to the representative cause event of the cluster.

## 2.5 Causality Prediction Rule Generation

The last phase of the learning is the creation of rules that will allow us, given a cause event, to generate a prediction about it. As the input cause event is matched against the node centroid, a naive approach would be to return the effect event of the matched centroid. This, however, would not provide us with the desired result. Assume an event $e_i$="Earthquake hits Haiti" occurred today, and that is matched to a node represented by the centroid: "Earthquake hits Turkey," whose effect is "Red Cross help sent to Ankara." Obviously, predicting that Red Cross help will be sent to Ankara because of an earthquake in Haiti is not reasonable. We would like to be able to abstract the relation between the past cause and past effect and learn a predicate clause that connects them, for example "Earthquake hits [Country Name]" yielding "Red Cross help sent to [Capital of Country]." During prediction, such a clause will be applied to the input event $e_i$, producing its predicted effect. In our example, the logical predicate clause would be *CapitalOf*, as *CapitalOf(Turkey)=* Ankara. When applied on the current event $e_i$, *CapitalOf(Haiti)* = Port-au-Prince, the output will now be "Red Cross help sent to Port-au-Prince." Notice that the the clauses can only be applied on certain types of objects – in our case, countries. The clauses can be of any length, e.g., the pair $\langle$"suspect arrested in Brooklyn," "Bloomberg declares emergency"$\rangle$ produces the clause *Mayor(BoroughOf(x))*, as Brooklyn is a borough of New York, whose mayor is Bloomberg.

We will now show how to learn such clauses for each node in the AT graph. Recall that the semantic network graph $G_O$ is an edge-labeled graph, where each edge is a triplet $\langle v_1, v_2, l \rangle$, where $l$ is a predicate (e.g., "CapitalOf"). The rule-learning procedure is divided into two main steps. First, we find an undirected path $p_i$ of length at most $k$ in $G_O$ between any object of the cause event to any object of the effect event. Note that we do not necessarily look for paths between two objects with the same role. In the above example, we found a path between the location of the cause event (Brooklyn) to the actor of the effect event (Bloomberg). Second, we construct a clause using the labels of the path $p_i$ as the predicates. We call this the *predicate projection* of size $k$, $pred = l_1, \ldots, l_k$ from an event $e_i$ to an event $e_j$. During prediction, the projections will be applied to the new event $e = \langle P^i, O_1, \ldots, O_4, t \rangle$ by finding an undirected path in $G_O$ from $O_i$ with the sequence of labels of $pred$. As $k$ is unknown, the algorithm, for each training example $\langle c_t, e_t \rangle$ in a node in the AT, finds all possible predicate paths with increasing sizes of $k$ from the objects of $c_t$ to the objects of $e_t$ in the $G_O$ graph. Each such path is weighted by the number of times it occurred in the node, the *support* of the rule. The full predicate generation procedure is described in Figure 3. The function *LearnPredicateClause* calls the inner function *FindPredicatePath* for different $k$ sizes and different objects from the given cause and effect events. *FindPredicatePath* is a recursive function that tries to find a path between the two objects in a graph of length $k$. It returns the labels of such a path if found. The rule generated is a template for generating the prediction of a future event given the cause event. An example of such a rule can be seen in Figure 4. Rules that return NULL are not displayed in the figure. In this example, when we generate object $O_1$ of the future event, we try to apply the path $\xrightarrow{l_1} \xrightarrow{l_2}$ on the object $O_4$ of the cause, thus generating possible objects that can be object $O_1$ of the prediction (see Section 2.6). Similarly, the path $\xrightarrow{l_1} \xrightarrow{l_2}$ is applied on $O_2$, generating more possible objects. For object $O_2$ of the prediction, a simple path of one label was generated. Therefore, during prediction, the possible objects for $O_2$ are the ones that connect to $O_4^{cause}$ with the label $l_8$ (if any). For object $O_3$ of the prediction, we use the $O_3^{cause}$. For $O_4$ no special rule was generated (*FindPredicatePath* returned NULL for all objects), and the final prediction will have $O_4^{effect}$.

## 2.6 Prediction

Given a trained model $\hat{g}$, it can be applied to a new event $e = \langle P^i, O_1, \ldots, O_4, t \rangle$ in order to produce its effects. The process is divided into two main steps – propagating the event in the AT to retrieve a set of matched nodes, and applying the rules of each matched node to produce the possible effects.

Given a new event, Pundit traverses the AT starting from the root. For each node in the search frontier, the algorithm computes the similarity ($SIM(e_i, e_j)$) of the input event to the centroid of each of the children on this node, and expands those children with better similarity than their parent. This idea can be stated intuitively as an attempt to find the nodes which are the least general but still similar to the new event. The full algorithm is illustrated in Figure 5. An illustration of the process can be seen in Figure 6. Here, an event of a bombing in Baghdad is received as input. The system searches for the least general cause event it has observed in the past (for simplicity we only show a short notation of the cause events in the AT). In our case, it is a generalized cluster: "bombing in city." Other candidates selected are the "military communication" cluster and the "bombing" cluster (as the node "bombing in worship area" has a lower score than "bombing").

For each node retrieved in the previous stage, its predicate projection, $pred$, is applied to the new event $e = \langle P^i, O_1, \ldots, O_4, t \rangle$. Informally, we can say that $pred$ is applied by finding an undirected path in $G_O$ from $O_i$ with the labels of $pred$. This rule generates a possible effect event from the retrieved node. The projection results are all the reached objects in the vertex. The formal explanation is that $pred$ can be applied if $\exists V_0 : O \subseteq V_0, \exists V_1 \ldots V_k : (V_0, V_1, l_1), \ldots (V_{k-1}, V_k, l_k) \in Edges(G_O)$. The projection results are all the

---

**Procedure** FindPredicatePath($curEntity, goalEntity, depth$)
   **If** $curEntity = goalEntity$ **Return** $\emptyset$
   **Else**
      **If** $depth = 0$ **Return** $NULL$
      **Else**
         **Foreach** $relation \in outEdges(curEntity)$
            $solution \leftarrow$ FindPredicatePath($relation(curEntity), goalEntity, depth - 1$)
            **If** $solution \neq NULL$
               **Foreach** $existingSolution \in Solutions$ :
                  **Return** $Solutions \bigcup (existingSolution||relation||solution)$
         **Return** $Solutions$

**Procedure** LearnPredicateClause($\langle P^c, O_1^c, \ldots, O_4^c, t^c \rangle, \langle P^e, O_1^e, \ldots, O_4^e, t^e \rangle, depth$)
   **Foreach** $O_i^c \in O^c, O_j^e \in O^e, k \in \{1 \ldots depth\}$
      $\text{rule}(j) \leftarrow \emptyset$
   **Foreach** $O_i^c \in O^c, O_j^e \in O^e, k \in \{1 \ldots depth\}$
      $\text{rule}(j) \leftarrow \text{rule}(j) \bigcup \{\langle O_i^c, \text{FindPredicatePath}(O_i^c, O_j^e, k)\rangle\}$
   **Return** rule

---

Figure 3: Procedure for generating a rule between two events by inferring paths between the two events in the causality graph.

objects $o \in V_k$. The projection results of all the nodes are weighted by the similarity of the target cause to the node $MGen$ and then ranked by the support of the rule (for tie breaking). If several $MGen$ exists, the highest similarity is considered. See Figure 7 for a complete formal description of the algorithm. In our example (Figure 6), the candidate "bombing in [city]" has the following rules:

1. $P^{effect} = happen$, $O_1^{effect} = riot$ , $O_4^{effect} = O_4^{cause}$

2. $P^{effect} = happen$, $O_1^{effect} = riot$ , $O_4^{effect} = O_4^{cause} \overset{main-street-of}{\longleftrightarrow}$

3. $P^{effect} = kill$, $O_2^{effect} = people$

4. $P^{effect} = condemn$, $O_1^{effect} = O_4 \overset{mayor-ofborough-of}{\longleftrightarrow}$, $O_2^{effect} = attack$

For clarity, for objects where no rule can be applied (the rule for the object is NULL), we use the effect concept of the matched training example.

For the event Baghdad Bombing ($O_1 = Bombing$, $O_4 = Baghdad$), applying the rules yields the following:

1. Baghdad Riots ($P^{effect} = happen$, $O_1^{effect} = riot$ , $O_4^{effect} = Baghdad$).

2. Caliphs Street Riots ($P^{effect} = happen$, $O_1^{effect} = riot$ , $O_4^{effect} = Caliphs\ Street$
   $\overset{main-street-of}{\longleftrightarrow} O_4^{cause}$).

$Rule(cause, effect) =$

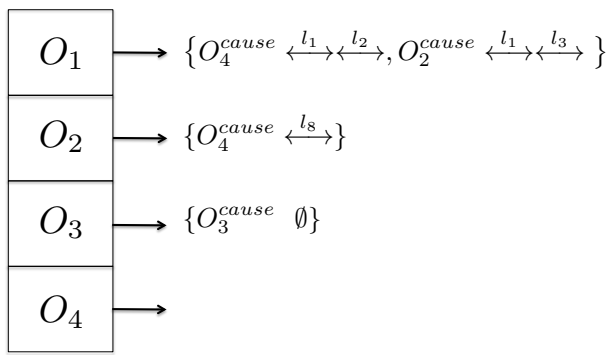

Figure 4: An example of a generated rule

3. People killed ($P^{effect} = kill$, $O_2^{effect} = people$).

4. This rule cannot be applied on the given event, as there is no outgoing edge of type borough-of for the node Baghdad.

## 2.7 Pruning Implausible Effects

In some cases, the system generated implausible predictions. For example, for the event ⟨lightning kills 5 people⟩, the system predicted that ⟨lightning will be arrested⟩. This prediction was based on generalized training examples in which people who killed other people got arrested, such as: ⟨*Man kills homeless man, man is arrested*⟩. But if we could determine how logical an event is, we could avoid such false predictions. In this section we discuss how we filter them out.

The goal of our filtering component is different from that of the predictor. While the predictor's goal is to generate predictions about future events, this component's goal is to monitor those predictions. While the predictor learns a causality relation between events, this component learns their plausibility.

The right way to perform such filtering is to utilize common sense knowledge for each action. Such knowledge would state the type of the actor and the object that can perform the action, the possible instruments with which the action can be preformed and the possible locations. If such knowledge would have existed, it would have identified that for the action arrest the object can be only human. However, such common sense knowledge is currently not available. Therefore, we had to resort to the common practice of using statistical methods.

---

**Procedure** PROPAGATION($e = \langle P^i, O_1, \ldots, O_4, t \rangle$)
   (1) $Candidates \leftarrow \{\}$
   (2) $Q \leftarrow$ new Queue
   (3) $Q.enqueue(G.root)$
   (4) **While** $Q \neq \emptyset$:
     (4.1) $cur \leftarrow Q.dequeue()$
     (4.2) **Foreach** $edge \in cur.OutEdges$:
        **If** $SIM(e, edge.Source) > SIM(e, edge.Destination)$:
        $Candidates \leftarrow Candidates \bigcup$
           $\{(edge.Source, SIM(e, edge.Source))\}$
        **Else** :
          $Q.enqueue(edge.Destination)$
   (5) **Return** $Candidates$

---

Figure 5: Procedure for locating candidates for prediction. The algorithm saves a set of possible matched results (Candidates), and a queue holding the search frontier (Q). In step 4, the algorithm traverses the graph. In step 4.2, for each edge, the algorithm tests whether the similarity of the new event $e$ to the parent node (edge.Source) is higher than to the child node (edge.Destination). If the test succeeds, the parent node, with its similarity score, is added to the possible results. After all edges are exposed, the algorithm returns the possible results in step 5.

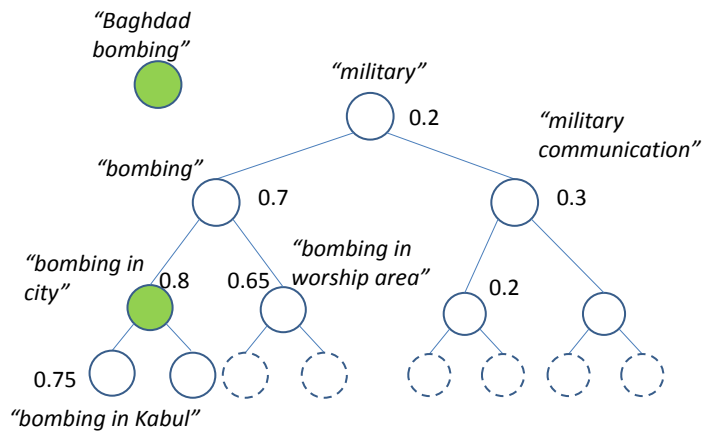

Figure 6: An event of a bombing in Baghdad is received as input. The system searches for the least general cause event it has observed in the past. In our case it is a generalized cluster: "bombing in city". The rule at this node now will be applied on the Baghdad bombing to generate the prediction.

---

**Procedure** FINDPREDICATEPATHOBJECTS($entity, path = \langle l_1 \ldots l_k \rangle$)
   (1) $Candidates \leftarrow \{\}$
   (2) $Q \leftarrow$ new Queue
   (3) $Q.enqueue(entity)$
   (4) $labelIndexInPath = 1$
   (5) **If** path.Count $== 0$: **Return** $\{entity\}$
   (6) **While** $Q \neq \emptyset$:
      $cur \leftarrow Q.dequeue()$
         **Foreach** $edge \in \{edge \in cur.OutEdges | edge.label = path[labelIndexInPath]\}$:
            **If** $labelIndexInPath = k$ :
               $Candidates \leftarrow Candidates \bigcup \{edge.Destination\}$
            **Else** :
               **If** $labelIndexInPath > k$: **Return** $Candidates$
               $Q.enqueue(edge.Destination)$
               $labelIndexInPath \leftarrow labelIndexInPath + 1$
   (7) **Return** $Candidates$

**Procedure** APPLYPREDICATECLAUSE($\langle P, O_1, \ldots, O_4, t \rangle, rule$)
   **Foreach** $i = 1 \ldots 4$
      $O_i^{prediction} \leftarrow \emptyset$
      **Foreach** $path = \{O_j, \{l_1 \ldots l_k\}\} \in rule(i)$
         $O_i^{prediction} \bigcup$ FindPredicatePathObjects($O_j, \langle l_1 \ldots l_k \rangle$)
   **Return** $\langle O_1^{prediction} \ldots O_4^{prediction} \rangle$

---

Figure 7: Procedure for applying a rule to a new given event. The main procedure is ApplyPredicateClause. This procedure generates the objects of the predicted event $O_1 \ldots O_4$ given a rule. The rule is a list of lists of tuples. Each tuple is a concept and a path. For each such tuple the function FindPredicatePathObjects is applied. This procedure finds objects that have a path whose labels connect to the given concept. Those objects are stored in Candidates (step 1). The algorithm holds a queue $Q$ with the frontier (step 2). The queue first holds the given entity (step 3). The procedure holds a counter indicating whether we followed the entire given path (step 4). The algorithm then checks whether there is an edge with the label path[labelIndexInPath] going out of the object at the head of the frontier. When the algorithm reaches the end of the given path ($labelIndexInPath = k$), it returns the candidates.

In the information extraction literature, identifying the relationship between facts and their plausibility has been widely studied. These methods usually estimate the prior probability of a relation by examining the frequency of its pattern in a large corpus, such as the Web (Banko et al., 2007). For example, for the relation $\langle$People,arrest,People$\rangle$ these methods return that this phrase was mentioned 188 times on the Web, and that the relation $\langle$People,arrest,[Natural Disaster]$\rangle$ was mentioned 0 times. Similarly, we estimate the prior probability of an event to occur from its prior appearance in the New York Times, our

---

**Procedure** Pruning Implausible Effects($ev = \langle P_i, O_1, \ldots, O_4, t \rangle, generalizationBound$)
   (1) **Foreach** $j \in 1 \ldots 4$:
     generalizationPath = {}
    **For** $i \in 0 \ldots generalizationBound$
       $Gen(O_i) \leftarrow FindPredicatePathObjects(O_j, generalizationPath)$
      generalizationPath $\leftarrow$ generalizationPath $\bigcup \{IsA\}$
   (2) **Return** $Average_{i,j,i \neq j}(Max_{o_1 \in Gen(O_i), o_2 \in Gen(O_j)} PMCI(o_1, o_2, i, j))$

---

Figure 8: A procedure for calculating the PMCI of an event. The procedure, at step 1, first generates all generalizations of type IsA of an object (with a path whose length is at most generalizationBound). For this purpose it uses the function FindPredicatePathObjects (defined in Figure 7). The generalization procedure is repeated on all objects comprising the event $ev$, and the result is stored in Gen. The final result of the algorithm is calculated in step 2. For two objects ($o_1, o_2$) in the generalization (Gen), which also contains the original objects, we find the maximum PMCI. We then compute the final result by averaging over this maximum PMCI.

primary source of news headlines. We then filter out events that, a priori, have very low probability to occur.

We present the algorithm in Figure 8. We calculated how many times the semantic concepts representing the event, or their immediate generalizations, actually occurred together in the past in the same semantic roles. In our example, we check how many times lightning or other natural phenomena were arrested. Formally, we define point-wise mutual concept information (PMCI) between two entities or verbs $o_i, o_j$ (e.g., lightning and arrest) in roles $r_i, r_j$ (e.g., actor and action) be defined as

$$PMCI(o_i, o_j, r_i, r_j) = \log \frac{p(o_i@r_i, o_j@r_j)}{p(o_i@r_i)p(o_j@r_j)}. \tag{3}$$

Given an event, we calculate the average PMCI of its components. The algorithm filters out predicted events that have low average PMCI. We assume that the cause and effect examples in the training are the ground truth, and should yield a high PMCI. Therefore, we evaluate the threshold for filtering from this training data. That is, we collected all the effects we observed in the training data and estimated their average PMCI on the entire NYT dataset.

The reader should note that applying such rules might create a problem. If in the past no earthquake occurred in Tokyo, the pruning procedure might return low plausibility. To handle these type of errors, we calculate the PMCI of the upper level categories of entities (e.g., natural disasters) rather than specific entities (e.g., earthquakes). We therefore restrict ourselves to only the two upper level categories.

## 3. Implementation Details

In the previous section, we presented a high-level algorithm that requires training examples $T$, knowledge about entities $G_O$, and event action classes $P$. One of the main challenges of this work was to build a scalable system to meet those requirements.

We present a system that mines news sources to extract events, constructs their canonical semantic model, and builds a causality graph on top of those events. The system crawled, for more than 4 months, several dynamic information sources (see Section 3.1 for details). The largest information source was the NYT archive, on which optical character recognition (OCR) was performed. The overall gathered data spans more than 150 consecutive years $(1851 - 2009)$.

For generalization of the objects, the system automatically reads Web content and extracts world knowledge. The knowledge was mined from structured and semi-structured publicly available information repositories. The generation of the causality graph was distributed over 20 machines, using a MapReduce framework. This process efficiently unites different sources, extracts events, and disambiguates entities. The resulting causality graph is composed of over 300 million entity nodes, one billion static edges (connecting the different objects encountered in the events), and over 7 million causality edges (connecting events that were found by Pundit to cause each other). Each rule in the AT was generated using an average of 3 instances with standard deviation of 2.

On top of the causality graph, a search and indexing infrastructure was built to enable search over millions of documents. This highly scalable index allows a fast walk on the graph of events, enabling efficient inference capabilities during the prediction phase of the algorithm.

### 3.1 World Knowledge Mining

The entity graph $G_o$ is composed of concepts from Wikipedia, ConceptNet (Liu & Singh, 2004), WordNet (Miller, 1995), Yago (Suchanek, Kasneci, & Weikum, 2007), and OpenCyc; the billion labeled edges of the graph $G_o$ are the predicates of those ontologies. In this section we describe the process by which this knowledge graph is created and the search system built upon it.

Our system creates the entity graph by collecting the above content, processing feeds, and processing formatted data sets (e.g., Wikipedia). Our crawler then archives those documents in raw format, and transforms them into RDF (Resource Description Framework) format (Lassila, Swick, Wide, & Consortium, 1998). The concepts are interlinked by humans as part of the Linked Data project (Bizer et al., 2009). The goal of Bizer et al.'s (2009) Linked Data project is to extend the Web by interlinking multiple datasets as RDF and by setting RDF links between data items from different data sources. Datasets include DBPedia (a structured representation of Wikipedia), WordNet, Yago, Freebase, and more. By September 2010 this had grown to 25 billion RDF triples, interlinked by around 395 million RDF links.

We use SPARQL queries as a way of searching over the knowledge graph. Experiments of the performance of those queries on the Berlin benchmark (Bizer & Schultz, 2009) provided evidence for the superiority of Virtuoso open source triple structures for our task.

## 3.2 Causality Event Mining and Extraction

Our supervised learning algorithm requires many learning examples to be able to generalize well. As the amount of temporal data is extremely large, spanning over millions of articles, the goal of obtaining human annotated examples becomes impossible. We therefore provide an automatic procedure to extract labeled examples for learning causality from dynamic content. In this work, we used the NYT archives for the years $1851 - 2009$, WikiNews, and the BBC – over 14 million articles in total (see data statistics in Table 1). Extracting causal relations between events in text is a hard task. The state-of-the-art precision of this task is around 37% (Do, Chan, & Roth, 2011). Our hypothesis is that most of the information regarding an event can be found in the headlines. These are more structured and therefore easier to analyze. Many times the headline itself can contain both the cause and effect event. We assume that only some of the headlines are describing events and developed an extraction algorithm to identify those headlines and to extract the events from them. News headlines are quite structured, and therefore the accuracy of this stage (performed on a representative subset of the data) is 78% (see Section 4.2.1). The system mines unstructured natural language text found in those headlines, and searches for causal grammatical patterns. We construct those patterns using *causality connectors* (Wolff, Song, & Driscoll, 2002; Levin & Hovav, 1994). In this work we used the following connectors:

1. Causal Connectives: the words *because*, *as*, and *after* as the connectors.

2. Causal prepositions: the words *due to* and *because of*.

3. Periphrastic causative verbs: the words *cause* and *lead to*.

We constructed a set of rules for extracting a causality pair. Each rule is structured as: ⟨Pattern, Constraint, Priority⟩, where Pattern is a regular expression containing a causality connector, Constraint is a syntactic constraint on the sentence on which the pattern can be applied, and Priority is the priority of the rule if several rules can be matched. The following constraints were composed:

1. Causal Connectives: The pattern [sentence1] after [sentence2] was used with the following constraints: [sentence1] cannot start with "when," "how," "where," [sentence2] cannot start with "all," "hours," "minutes," "years," "long," "decades." In the pattern "After [sentence1], [sentence2]" we add the constraint that [sentence1] cannot start with a number. This pattern can match the sentence "after Afghan vote, complaints of fraud surface" but will not match the sentence "after 10 years in Lansing, state lawmaker Tom George returns". The pattern "[sentence1] as [sentence2]" was used with the constraint of [sentence2] having a verb. Using the constraint, the pattern can match the sentence "Nokia to cut jobs as it tries to catch up to rivals", but not the sentence "civil rights photographer unmasked as informer."

2. Causal prepositions: The pattern [sentence1]["because of," "due to"] [sentence2] only required constraints that [sentence1] does not start with "when," "how," "where."

3. Periphrastic causative verbs: The pattern [sentence1] ["leads to," "Leads to," "lead to," "Lead to," "led to," "Led to"] [sentence2] is used, where [sentence1] cannot con-tain "when," "how," "where," and the prefix cannot be "study" or "studies." Additionally, as we consider periphrastic causative *verbs*, we do not allow additional verbs in [sentence1] or [sentence2].

The result of a rule application is a pair of sentences – one tagged as a cause, and one tagged as an effect.

Given a natural-language sentence (extracted from an article headline), representing an event (either during learning or prediction), the following procedure transforms it into a structured event:

1. Root forms of inflected words are extracted using a morphological analyzer derived from WordNet (Miller, 1995) stemmer. For example, in the article headline from 10/02/2010: "U.S. attacks kill 17 militants in Pakistan", the words "attacks," "killed" and "militants" are transformed to "attack," "kill," and "militant" respectively.

2. Part-Of-Speech tagging (Marneffe, MacCartney, & Manning, 2006) is performed, and the verb is identified. The class of the verb is identified using the VerbNet vocabulary (Kipper, 2006), e.g., kill belongs to $P$ =murder class.

3. A syntactic template matching the verb is applied to extract the semantic relations and thus the roles of the words (see example in Figure 10). Those templates are based on VerbNet, which supplies for each verb class a set of syntactic templates. These templates match the syntax to the thematic roles of the entities in the sentence. We match the templates even if they are not continuous in the sentence tree. This allows the match of a sentence even where there is an auxiliary verb between the subject and the main transitive verb. In our example, the template is "NP1 V NP2," which transforms NP1 to "Agent", and NP2 to "Patient." Therefore, we match U.S. attacks to be the *Actor*, and the militant to be the *Patient* . If no template can be matched, the sentence is transformed into a typed-dependency graph of grammatical relations (Marneffe et al., 2006). In the example, U.S. attacks is identified as the subject of the sentence (candidate for Actor), militants as the object (candidate for Patient), and Pakistan as the preposition (using Locations lexicons). Using this analysis, we identify that the *Location* is Pakistan.

4. Each word in $O_i$ is mapped to a Wikipedia-based concept. If a word matches more than one concept, we perform disambiguation by computing the cosine similarity between the body of the news article and the body of the Wikipedia article associated with the concept. For example, U.S was matched to several concepts, such as United States, University of Salford, and Us (Brother Ali album). The most similar by content was the Wikipedia concept United States. If a word in $O_i$ is not found in Wikipedia, it is treated as a constant, i.e., generalization will not be applied on it, but it will be used during similarity calculation. That is, $dist_{Gen}(const_1, const_2) = 0$ if $const_1 = const_2$, or $dist_{Gen}(const_1, const_2) = k$ otherwise. In our experiments, we set $k = 4$, as it was the length of the longest distance found between two concepts in $G_O$.

5. The time of the event $t$ is the time of the publication of the article in the news, e.g., $t$ =10/02/2010.

| Data Source | Number of Titles Extracted |
|-------------|----------------------------|
| NYT         | 14,279,387                 |
| BBC         | 120,445                    |
| WikiNews    | 25,808                     |

Table 1: Data Summary.

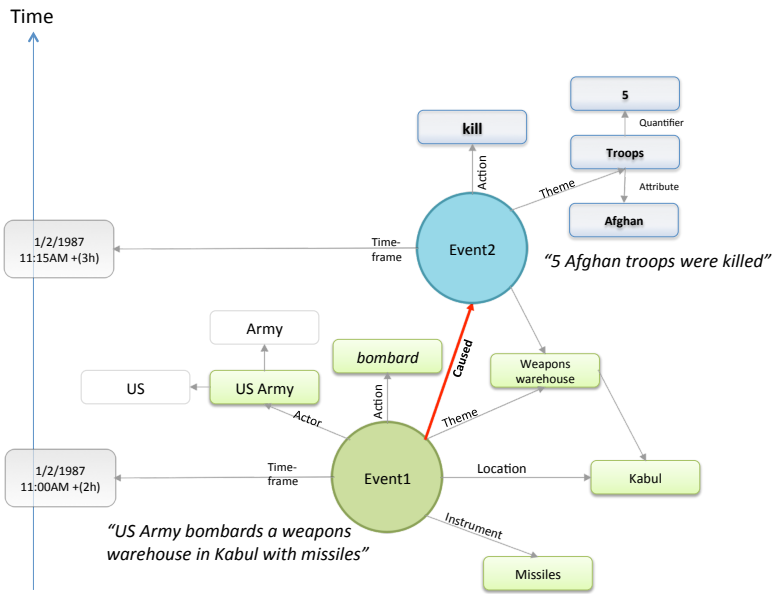

Figure 9: A pair of events in the causality graph. The first represents a cause event and the second represents the effect event. Both were extracted from the headline published on 1/2/1987: 5 Afghan troops killed after US army bombards warehouse in Kabul.

In our example, the final result is the event $e = \langle$Murder-Class, United States of America, Militant, NULL, Pakistan, 10/02/2010$\rangle$ . The final result of this stage is a causality graph composed of causality event pairs. Those events are structured as described in Section 2.1. We illustrate such a pair in Figure 9.

In certain cases, additional heuristics were needed in order to deal with the brevity of news language. We used the following heuristics:

1. Missing Context – In "McDonald's recalls glasses due to cadmium traces," the extracted event "cadmium traces" needs additional context – "Cadmium traces [in McDonald's glasses]." If an object is missing, the first sentence ([sentence1]) subject is used.

| Class Hit-18.1 | | |
|---|---|---|
| Roles and Restrictions: | | |
| Agent[int control] Patient[concrete] Instrument[concrete] | | |
| Members: bang, bash, hit, kick, . . . | | |
| Frames: | | |
| Example | Syntax | Semantics |
| Paula hit the ball | Agent V Patient | cause(Agent, E)<br>manner(during(E),<br>    directedmotion, Agent)<br>!contact(during(E),<br>    Agent, Patient)<br>manner(end(E),forceful,<br>    Agent)<br>contact(end(E), Agent,<br>    Patient) |

Figure 10: VerbNet Template.

2. Missing entities and verbs – the text "22 dead" should be structured to the event "22 [people] [are] dead." If a number appears as the subject, the word people is added and used as the subject, and "be" is added as the verb.

3. Anaphora resolution – the text "boy hangs himself after he sees reports of Hussein's execution" is modeled as "$[boy_1]$ sees reports of Hussein's execution" causes "$[boy_1]$ hangs $[boy_1]$" (Lappin & Leass, 1994).

4. Negation – the text "Matsui is still playing despite his struggles" should be modeled as: "[Matsui] struggles" causes the event "Matsui is [not] playing". Modeling preventive connectors (e.g., despite) requires negation of the modeled event.

## 4. Experimental Evaluation

In this section, we describe a set of experiments performed to evaluate the ability of our algorithms to predict causality. We first evaluate the predictive precision of our algorithm, continue with analyzing each part of the algorithm separately, and conclude with a qualitative evaluation.

### 4.1 Prediction Evaluation

The prediction algorithm was trained using news articles from the period $1851 - 2009$. The world knowledge used by the algorithm was based on Web resource snapshots (Section 3) dated until 2009. The evaluation was performed on separate data – Wikinews articles from the year 2010. We refer to this data as the *test data*.

As the task tackled by our algorithm has not been addressed before, we could not find any baseline algorithm to compare against. We therefore decided to compare our algorithm's performance to that of human predictors. Our algorithm and its human competitors were assigned the basic task of predicting what event a given event might cause. We evaluate each such prediction using two metrics. The first metric is *accuracy*: whether the predicted event actually occurred in the real world. There are two possible problems with this metric. First, a predicted event, though plausible, still might not actually have occurred in the real world. Second, the predicted event might have happened in the real world but was not caused by the given event, for example, in trivial predictions that are always true ("the sun will rise"). We therefore use an additional metric, event *quality*, the likelihood that the predicted event was caused by the given event.

The experiments were conducted as follows:

1. Event identification – our algorithm assumes that the input to the predictor $h$ is an event. To find news headlines that represent an event, we randomly sample $n = 1500$ headlines from the test data. For each headline, a human evaluator is requested to decide whether the headline is an event that can cause other events. We denote the set of headlines labeled as events as $E$. We again randomly sample $k = 50$ headlines from $E$. We denote this group as $C$.

2. Algorithm event prediction – on each headline $c_i \in C$, Pundit performs event extraction, and produces an event $Pundit(c_i)$ with the highest score of being caused by the event represented by $c_i$. Although the system provides a ranked list of results, to simplify the human evaluation of theses results, we consider only the highest score prediction. If there is a tie for the top score, we pick one at random. The results of this stage are the pairs: $\{(c_i, Pundit(c_i)) | c_i \in C\}$.

3. Human event prediction – For each event $c_i \in C$, a human evaluator is asked to predict what that event might cause. Each evaluator is instructed to read a given headline and predict its most likely outcome, using any online resource and with no time limit. The evaluators are presented with empty structured forms with the 5 fields for the output event they need to provide. The human result is denoted as $human(c_i)$. The results of this stage are the pairs: $\{(c_i, human(c_i)) | c_i \in C\}$.

4. Human evaluation of the results –

   (a) Quality: We present $m = 10$ people with a triplet $(c_i, human(c_i), Pundit(c_i))$. The human evaluators are asked to grade $(c_i, human(c_i))$ and $(c_i, Pundit(c_i))$ on a scale of 0-4 (0 is a highly implausible prediction and 4 is a highly plausible prediction). They were allowed to use any resource and were not limited by time. The human evaluators were different from those who performed the predictions.

   (b) Accuracy: For each predicted event, we checked the news (and other Web resources), up to a year after the time of the cause event, to see whether the predicted events were reported.

Human evaluation was conducted using Amazon Mechanical Turk, an emerging utility for performing user study evaluations, which was shown to be very precise for certain tasks (Kittur, Chi, & Suh, 2008). During the evaluation, tasks are created by routing a question to random users and obtaining their answers. We filtered the raters using a CAPTCHA. We restricted to only US-based users, as the events used by our system are extracted from the NYT. We did not perform any other manual filtering of the results. The average times for all human tasks are reported in table 2. We observed that the most time-consuming task for humans was to verify that the event indeed happened in the past. The other time-consuming task was Human Event Prediction. This is not surprising, as both cases required more use of external resources, whereas the quality evaluation only measured whether those events make sense. Additionally, we manually investigated the human evaluations in each category, and did not find correlation between the response time and quality of the human prediction. As we used Mechanical Turk, we do not know which external resources the evaluators used. We measured inter-rater reliability using Fleiss' kappa statistical test, where $\kappa$ measures the consistency of the ratings. For the raters in our test, we obtained $\kappa = 0.3$, which indicates fair agreement (Landis & Koch, 1977; Viera & Garrett, 2005). This result is quite significant, for the following reasons:

1. Conservativeness of this measure.

2. Subjectivity of the predictions – asking people whether a prediction makes sense often leads to high variance in responses.

3. Small dataset – the tests were performed with 10 people asking to categorize into 5 different scales of plausibility over 50 examples.

4. Lack of formal guidelines for evaluating the plausibility of a prediction – no instructions were given to the human evaluators regarding what should be considered plausible and what is not.

| Human Event Identification | Human Event Prediction | Human Evaluation (Quality) | Human Evaluation (Accuracy) |
|---|---|---|---|
| 1 min 26 sec | 4 min 10 sec | 1 min 44 sec | 6 min 24 sec |

Table 2: Response times of human evaluators for the different evaluation tasks.

|  | [0-1) | [1-2) | [2-3) | [3-4] | Average Quality |
|---|---|---|---|---|---|
| Pundit | 0 | 2 | 19 | 29 | **3.08** |
| Humans | 0 | 3 | 24 | 23 | 2.86 |

Table 3: Quality results. The histogram of the rankings of the users for humans and the algorithm.

Additionally, for comparison, similar tasks in natural language processing, such as sentence formality identification (Lahiri, Mitra, & Lu, 2011), usually reach kappa values of $0.1 - 0.2$.

The quality evaluation yielded that Pundit's average predictive precision is $3.08/4$ (3 is a "plausible prediction"), as compared to $2.86/4$ for the humans. For each event, we average the results of the $m$ rankers, producing an average score for the algorithm's performance on the event, and an averaged score for the human predictors (see Table 3). We performed a paired t-test on the $k$ paired scores. The advantage of the algorithm over the human evaluators was found to be statistically significant, with $p \leq 0.05$.

| Algorithm | Average Accuracy |
|-----------|------------------|
| Pundit    | **63%**          |
| Humans    | 42%              |

Table 4: Prediction accuracy for both human and algorithm.

The accuracy results are reported in Table 4. We performed a Fisher's exact test (as the results are binary) on the $k$ paired scores. The results were found to be statistically significant, $p \leq 0.05$.

## 4.2 Component Analysis

In this section, we report the results of our empirical analysis of the different parts of the algorithm.

### 4.2.1 Evaluation of the Extraction Process

In Section 3.1, we described a process for extracting causality pairs from the news. These pairs are used as a training set for the learning algorithm. This process consists of two main parts: causality identification and event extraction. We perform a set of experiments to provide insights on this extracted training data quality.

**Causality Extraction Experiment**   The first step in building a training set consists of using causality patterns to extract pairs of sentences for which the causality relation holds. To assess the quality of this process, we randomly sampled 500 such pairs from the training set and presented them to human evaluators. Each pair was evaluated by 5 humans. We filtered the raters using a CAPTCHA and filtered out outliers. The evaluators were shown two sentences the system believed to be causally related and they were asked to evaluate the plausibility of this relation on a scale of 0-4.

The results show that the averaged precision of the extracted causality events is 3.11 out of 4 (**78%**), where 3 means a plausible causality relation, and 4 means a highly plausible causality relation. For example, the causality pair: "pulling over a car" → "2 New Jersey police officers shot," got a very high causality precision score, as this is a plausible cause-effect relation, which the system extracted from the headline "2 New Jersey Police Officers Shot After Pulling Over a Car."

For comparison, other temporal rule extraction systems (Chambers, Wang, & Jurafsky, 2007) reach precision of about 60%. The better performance of our system can be explained by our use of specially crafted templates (we did not attempt to solve the general problem of temporal information extraction).

Most causality pairs extracted were judged to be of high quality. The main reason for errors was that some events, although reported in the news and matching the templates we have described, are not common-sense causality knowledge. For example, "Aborted landing in Beirut" → "Hijackers fly airliner to Cyprus", was rated unlikely to be causally related, although the event took place on April 09, 1988.

|  | Action | Actor | Object | Instrument | Location | Time |
|---|---|---|---|---|---|---|
| Quality Precision | 93% | 74% | 76% | 79% | 79% | 100% |

Table 5: Extraction precision for each of the 5 event components using the causality patterns.

| Actor Matching | Object Matching | Instrument Matching | Location Matching | Action Matching |
|---|---|---|---|---|
| 84% | 83% | 79% | 89% | 97% |

Table 6: Entity-to-ontology matching precision.

**Event Extraction Experiment**   After a pair of sentences is determined to have a casualty relation, our algorithm extracts a structured event from each of the sentences. This event includes the following roles: *action*, *actor*, *object*, *instrument*, and *time*.

To assess the the quality of this process, we used the 500 pairs from the previous experiment and presented each of the 1000 associated sentences to 5 human evaluators. The evaluators were shown a sentence together with its extracted roles: *action*, *actor*, *object*, *instrument*, and *time*, and they were asked to mark each role assignment as being right or wrong.

Table 5 shows that the precision for the extracted event components ranges from $74 - 100\%$. In comparison, other works (Chambers & Jurafsky, 2011) for extracting entities for different types of relations reach $42 - 53\%$ precision. The higher precision of our results is mainly due to the use of domain-specific templates.

We performed additional experiments to evaluate the matching of every entity from the above experiment to the world-knowledge ontology. The matching was based on semantic similarity. Each ranker was asked to indicate whether the extracted entity was mapped correctly to a Wikipedia URI. The results are summarized in Table 6.

### 4.2.2 Evaluation of the Event Similarity Algorithm

Both the learning and prediction algorithms strongly rely on the event similarity function *dist* described in Section 2.4. To evaluate the quality of this function, we randomly sampled 30 events from the training data and found for each the most similar event from the entire past data (according to the similarity function). A human evaluator was then asked to evaluate the similarity of these events on a scale of 1–5. We repeated the experiment, replacing the average aggregator function $f$ with minimum and maximum functions.

The results are presented in Table 7. The general precision of the average function was high (3.9). Additionally, the average function performed substantially (confirmed by a t-test) better than over the minimum and maximum. This result indicates that distance functions that aggregate over several objects of the structured event (rather than just selecting the minimum or maximum of one of the events) yield the highest performance.

| Minimum | Maximum | Average |
|---------|---------|---------|
| 1.9 | 3.5 | 3.9 |

Table 7: Comparison of the different aggregations for the event-similarity $f$.

### 4.2.3 The Importance of Abstraction

Given a cause event whose effect we wish to predict, we use the algorithm described in Section 2.4 to identify similar generalized events. To evaluate the importance of this stage, we compose an alternative matching algorithm, similar to the nearest-neighbor approach (as applied by the work by Gerber, Gordon, & Sagae, 2010), that matches the cause event to the cause events of the training data. Instead of building an abstraction tree, the algorithm simply finds the closest cause in the past based on text similarity. We then rank the matched results using TF-IDF measure.

We applied both our original algorithm and this baseline algorithm on the 50 events used for prediction. For each event, we asked a human evaluator to compare the prediction of the original and the baseline algorithm. The results showed that in **83%** of the cases the predictions with generalization were rated as more plausible than those of the nearest-neighbor approach without generalization.

### 4.2.4 Analysis of Rule Generation Application

In order to generate an appropriate prediction with respect to the given cause event, a learned rule is applied, as described in Section 2.5. We observe that in **31%** of the predictions, a non-trivial rule was generated and applied (that is, a non-NULL rule that does not simply output the effect it observed in the matched past cause-effect pair example). Out of those, the application predicted correctly in more than **90%** of the cases and generated a plausible object in the effect. These results indicate that generalization and rule-generation techniques are essential to the performance of the algorithm.

### 4.2.5 Analysis of Pruning Implausible Causation

To eliminate situations in which a generated prediction is implausible, we devised an algorithm (Section 2.7) that prevents implausible predictions. We randomly selected 200 predictions from the algorithm predictions based on the human-labeled events extracted from the Wikinews articles (see Section 4.1). A human rater was requested to label predictions that are considered implausible. We then applied our filtering rules on the 200 predictions as well. The algorithm found 15% of the predictions to be implausible with 70% precision and 90% recall with respect to the human label. A qualitative example of a filtered prediction is "Explosion will surrender" for the cause event "Explosion in Afghanistan kills two."

## 4.3 Qualitative Analysis

For a better understanding of the algorithm's strengths and weaknesses we now present some examples of results. Given the event "Louisiana flood," the algorithm predicted that [number] people will flee. The prediction process is illustrated in Figure 11.

1. *Raw data*:

   The above prediction was based on the following raw news articles:

   (a) 150000 flee **as** hurricane nears North Carolina coast.

   (b) A million flee **as** huge storm hits Texas coast.

   (c) Thousands flee **as** storm whips coast of Florida.

   (d) Thousands in Dallas Flee Flood **as** Severe Storms Move Southwest.

2. *Causality pair extraction*:

   The "as" template was used to process the above headlines into the following structured events:

   (a) *Cause Event*: near (Action); hurricane (Actor); Coast(Object); North Carolina (Object Attribute) ; (Instrument); Carolina (Location); 31 Aug 1993 (Time).
   *Effect Event*: flee (Action); People (Actor); 150000(Actor Attributes); Carolina (Location); 31 Aug 1993 (Time).

   (b) *Cause Event*: hit (Action); Storm (Actor); Huge (Actor Attributes); Coast(Object); Texas (Object Attribute); Texas (Location); 13 Sep 2008 (Time).
   *Effect Event*: flee (Action); People (Actor); million(Actor Attributes); Texas (Location); 13 Sep 2008 (Time).

   (c) *Cause Event*: whip (Action); Storm (Actor); Coast(Object); Florida (Object Attribute); Florida (Location); March 19, 1936 (Time).
   *Effect Event*: flee (Action); People (Actor); thousands(Actor Attributes); Florida (Location); March 19, 1936 (Time).

   (d) *Cause Event*: move (Action); Storm (Actor); Severe (Actor Attributes); Dallas (Location); May 27, 1957 (Time).
   *Effect Event*: flee (Action); People (Actor); thousands(Actor Attributes); Flood(Object); Dallas (Location); May 27, 1957 (Time).

3. *Learning the abstraction tree*:

   The above four events were clustered together in the AT. They were clustered in the same node because the causes were found to be similar: the actors were all weather hazards and the location was a state of the United States. The effects were found to be similar as the actions and actors were similar across all events, and the actor attributes were all numbers. For this generalization, the following world knowledge was used:

   (a) Storm, hurricane and flood are "weather hazards" (extracted from the in-category relation in Wikipedia).

   (b) Carolina, Texas, and California are located in the "United States" (extracted from the located-in relation in Yago).

4. *Prediction*:

During the prediction, the event "Louisiana flood" (which did not occur in the training examples) was found most similar to the above node, and the node rule output was that [number] people will flee.

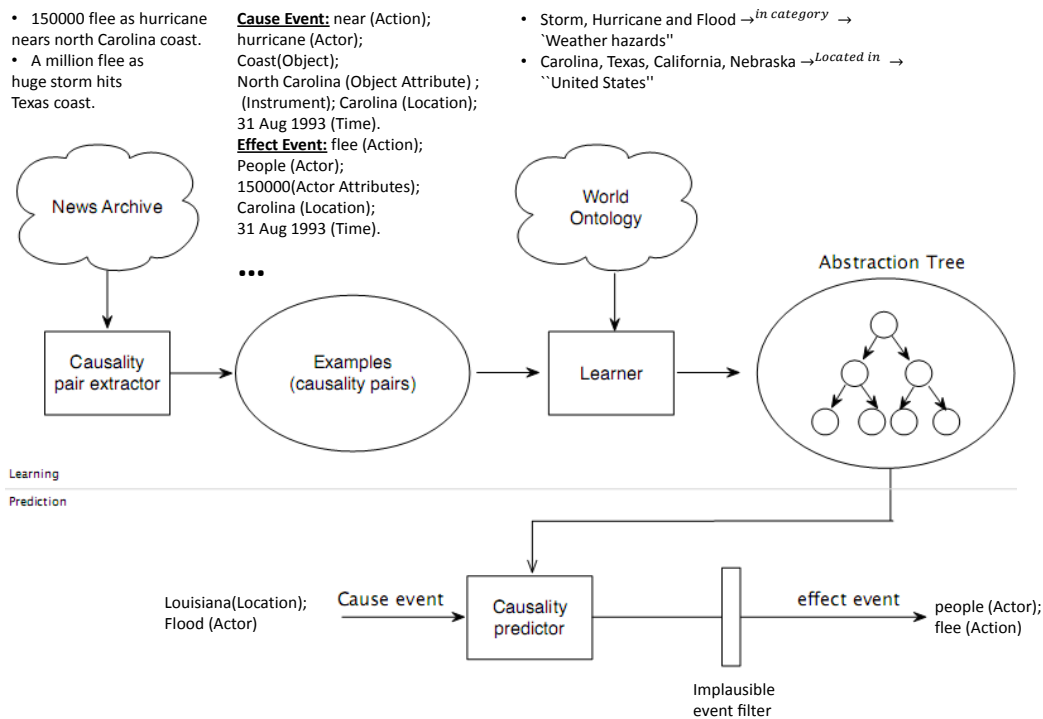

Figure 11: Examples of a prediction

As another example, given the event "6.1 magnitude aftershock earthquake hits Haiti," the highest matching predictions were: "[number] people will be dead," "[number] people will be missing," "[number] magnitude aftershock earthquake will strike island near Haiti" and "earthquake will turn to United States Virgin Islands." The first three predictions seem very reasonable. For example, the third prediction came from a rule that natural disasters hitting countries next to a shore tend to affect nearby countries. In our case it predicted that the earthquake will affect the United States Virgin Islands, which are geographically close to Haiti. The fourth prediction, however, is not very realistic as an earthquake cannot change its course. It was created from a match with a past example of a tornado hitting a country on a coast. The implausible causation filters this prediction, as it has very low PMCI, and the output of the system is "[number] people will be dead". This example is also interesting, as it issues a prediction using spatial locality (the United States Virgin Islands are [near] Haiti).

Additional examples out of the 50 in the test and their predictions can be seen in Table 8.

| Cause event | Human-predicted effect event | Algorithm-predicted effect event |
|---|---|---|
| Al-Qaida demands hostage exchange | **Al-Qaida exchanges hostage** | A country will refuse the demand |
| Afghanistan's parliament rejects Karzai's cabinet nominations | **Parliament accepts Karzai's cabinet nominations** | **Many critics of rejection** |
| Remains of 1912 expedition plane found in Antarctica | Europe museums vie for remains | **Enduring mystery will be solved in Antarctica** |
| North Korea seeks diplomatic relations with the US | UN officials offer mediation services | **North Korea rift will grow** |
| Volcano erupts in Democratic Republic of Congo | Scientists in Republic of Congo investigate lava beds | **Thousands of people flee from Congo** |
| Iceland's President vetoes repayment of Ice save losses | **Banks in Reykjavik report record withdrawals** | **Official administration reaction issued** |
| Death toll from Brazil mudslides rises to sixty | **Rescuers in Brazil abandon rescue efforts** | **Testimonies will be heard** |
| 7.0 magnitude earthquake strikes Haitian coast | **Tsunami in Haiti affects coast** | **Tsunami warning is issued** |
| 2 Palestinians reportedly shot dead by Israeli troops | Israeli citizens protest against Palestinian leaders | **Israeli troops will face scrutiny** |
| Professor of Tehran University killed in bombing | Tehran students remember slain professor in memorial service | **Professor funeral will be held** |
| Alleged drug kingpin arrested in Mexico | Mafia kills people with guns in town | **Kingpin will be sent to prison** |
| UK bans Islamist group | Islamist group would adopt another name in the UK | **Group will grow** |
| China overtakes Germany as world's biggest exporter | German officials suspend tariffs | **Wheat price will fall** |
| Cocaine found at Kennedy Space Center | Someone will be fired | People will be arrested |

Table 8: Human and algorithm predictions for events. Predictions in bold were labeled by the evaluators as correct predictions.

## 4.4 Discussion

In our experiments we only report the precision of our algorithms. Further experiments measuring the recall of the system are necessary. However, in our experiments each validation step required human intervention. For example, validating that a prediction occurred in the future news. In order to perform a full recall experiment one should apply the algorithm on all the news headlines reported on a certain day and measure the appearance of all the corresponding predictions in the future news. Unfortunately, performing human validation on such a large prediction space is hard. We leave the task of performing experiments to provide a rough estimate of recall to future work.

It is common practice to compare system performance to previous systems tackling the same problem. However, the ambitious task we tackled in this work had no immediate baselines to compare with. That is, there was no comparable system neither in scale nor in the ability to take an arbitrary cause event in natural language and output an effect event in natural language. Instead, we compared to the only agents we know capable of performing such a task – humans.

Although the results indicate the superiority of the system over such human agents, we do no claim that the system predictions perform better than humans. We rather provide evidence that the system provides similar predictions to that of humans, and sometimes even outperforms human ability to predict, as can be supported by the superiority of the system in the accuracy evaluation.

To fully support the claim of superiority of the system over humans, wider experiments should be performed. Experiments larger by an order of magnitude can provide results with higher agreement between raters and shed light on the different types of events where the system's performance is better. Additionally, more experiments comparing the system performance to that of experts in the fields of each individual prediction can be valuable as well. At this point, we assume the performance of experts would be higher than that of our algorithm. The main reason for this is the causality knowledge used to train the algorithms. This knowledge is extracted from headlines that tend to have simple causality contents, which is easily understandable by the general population. This type of knowledge limits the complexity of the predictions that can be made by Pundit. Pundit predictions therefore that tend to be closer to common knowledge of the average human. In order to predict more complex events we would need to rely on better training examples than news headlines alone.

The evaluation presented in this section provides evidence of the quality of the predictions that the system can provide. Our results are impressive in the sense that they are comparable to that of humans, thus providing evidence to the ability of a machine to perform one of the most desirable goals of general AI.

## 5. Related Work

We are not aware of any work that attempts to perform the task we face: receive arbitrary news events in natural language representation and predict events they can cause. Several works, however, deal with related tasks. In general, our work does not focus on better information extraction or causality extraction techniques, but rather on how this information can be leveraged for prediction. We present novel methods of combining world knowledge with event extraction methods to represent coherent events, and present novel methods for rule extraction and generalization using this knowledge.

### 5.1 Prediction from Web Behavior, Books and Social Media

Several works have focused on using search-engine queries for prediction in both traditional media (Radinsky, Davidovich, & Markovitch, 2008) and blogs (Adar, Weld, Bershad, &

Gribble, 2007). Ginsberg et al. (2009) used queries for predicting H1N1 influenza outbreaks. In the context of causality recognition, Gordon, Bejan, and Sagae (2011) present a methodology for mining blogs to extract common-sense causality. The evaluation is done on a human-labeled dataset where each test consists of a fact and two possible effects. Applying point-mutual information to personal blog stories, the authors select the best prediction candidate. The work differs from ours in that the authors focus on personal common-sense mining and do not consider whether their predictions actually occurred. Other works focused on predicting Web content change itself. For example, Kleinberg (2002, 2006) developed general techniques for summarizing the temporal dynamics of textual content and for identifying bursts of terms within content. Similarly, Amodeo, Blanco, and Brefeld (2011) built a time series model over publication dates of documents relevant to a query in order to predict future bursts. Social media were used to predict riots (Kalev, 2011) and movie box office sales (Asur & Huberman, 2010; Joshi, Das, Gimpel, & Smith, 2010; Mishne, 2006). Other works (Jatowt & Yeung, 2011; Yeung & Jatowt, 2011; Michel, Shen, Aiden, Veres, Gray, Google Books Team, Pickett, Hoiberg, Clancy, Norvig, Orwant, Pinker, Nowak, & Aiden, 2011) have explored the use of text mining techniques over news and books to explain how culture develops, and what people's expectations and memories are.

Our work differs from the above in several ways: First, we present a *general-purpose* prediction algorithm rather than a domain-specific one. Second, unlike the above works, ours combines a variety of *heterogenous online sources*, including world knowledge mined from the Web. Finally, we focus on generation of future event predictions represented entirely in natural language, and provide techniques to enrich and generalize historical events for the purpose of future event prediction.

## 5.2 Textual Entailment

A related topic to our work is that of textual entailment (TE) (Glickman, Dagan, & Koppel, 2005). A text $t$ is said to entail a textual hypothesis $h$ if people reading it agree that the meaning of $t$ implies the truth of $h$. TE can be divided into three main categories: recognition, generation, and extraction. In this section, we provide a short summary of the first two categories. For a more detailed overview we refer the reader to the survey by Androutsopoulos and Malakasiotis (2010). We then discuss the specific task of causality extraction from text in Section 5.3.4.

### 5.2.1 Textual Entailment recognition

In this task, pairs of texts are given as input, and the output is whether TE relations hold for the pair. Some approaches map the text to logical expressions (with some semantic enrichment, using WordNet, for example) and perform logical entailment checks, usually using theorem provers (Raina, Ng, & Manning, 2005; Bos & Markert, 2005; Tatu & Moldovan, 2005). Other approaches map the two texts to a vector space model, where each word is mapped to strongly co-occurring words in the corpus (Mitchell & Lapata, 2008), and then similarity measures over those vectors are applied. Some measure syntactic similarity by applying graph similarity measure on the syntactic dependency graphs of the two texts (Zanzotto, Pennacchiotti, & Moschitti, 2009). Similarly, other methods measure the semantic distance similarity between the words in text (Haghighi, 2005), usually exploiting other resources such as WordNet as well. The last set of approaches represents the two texts in a single feature vector and trains a machine learning algorithm, which later, given two new texts represented via a vector, can determine whether they entail each other (Bos & Markert, 2005; Burchardt, Pennacchiotti, Thater, & Pinkal, 2009; Hickl, 2008). For example, Glickman et al. (2005) show a naive Bayes classifier trained on lexical features, i.e., the number of times that words of $t$ appeared with words of $h$. Other features usually include polarity (Haghighi, 2005), whether the theorem prover managed to prove entailment (Bos & Markert, 2005), or tagging of the named entities to the categories people, organizations, or locations.

### 5.2.2 Textual Entailment Generation

Here we discuss TE generation, where, given an expression, the system should output a set of expressions that are entailed by the input. This task is most closely related to the one presented in this work: in TE generation, a text is received and an entailed text is generated as output. Androutsopoulos and Malakasiotis (2010) mention that no benchmarks exist to evaluate this task, and the most common and costly approach is to evaluate using human judges. We also encountered this difficulty in our own task, and performed human evaluation.

TE generation methods can be divided into two types: those that use machine translation techniques and those that use template-based techniques. Those that use machine translation techniques try to calculate the set of transformations with the highest probability, using a training corpus. Quirk, Brockett, and Dolan (2004) cluster news articles referring to the same event, select pairs of similar sentences, and apply the aforementioned techniques. Other methods use template-based approaches on large corpora, such as the Web. Some methods (Idan, Tanev, & Dagan, 2004) start with an initial seed of sentences (composed of entities), and use a search engine to find other entities for which these entailment relations hold. Those relations are used as templates. To find additional entities for which these relations hold, the relations themselves are then searched again. The TE generation system, given a text, matches it to a template and outputs all the texts that matched this template. Others (Ravichandran, Ittycheriah, & Roukos, 2003) also add additional filtering techniques on those templates.

Our work is most closely related to the template-based approach. We have crafted a new set of templates to extract causality pairs from the news.

### 5.3 Information Extraction

Information Extraction is the study of automatic extraction of information from unstructured sources. We categorizes the types of information extracted into three types: entities, relationships between entities, and higher-order structures such as tables and lists. The most closely related tasks to ours are those of entity extraction and relation extraction; for the rest we refer the reader to the survey by Sarawagi (2008). The former task, similar to our process of extracting concepts, deals with extracting noun phrases from text. In the latter task, given a document and a relation as input, the problem is to extract all entity pairs in the document for which this relation holds. Whereas the above works deal only with one element of our problem – extraction of information needed to understand a given causality, we deal with the actual causality prediction. We do not claim to create more precise information extraction methods, but rather try to leverage all this knowledge to perform an important AI task – future event prediction.

### 5.3.1 Entity Extraction

For entity extraction, two categories of methods exist – rule-based and statistical methods. Rule-based methods (Riloff, 1993; Riloff & Jones, 1999; Jayram, Krishnamurthy, Raghavan, Vaithyanathan, & Zhu, 2006; Shen, Doan, Naughton, & Ramakrishnan, 2007; Ciravegna, 2001; Maynard, Tablan, Ursu, Cunningham, & Wilks, 2001; Hobbs, Bear, Israel, & Tyson, 1993) define contextual patterns consisting a regular expression over features of the entities in the text (e.g., the entity word, part-of-speech tagging). Those rules are either manually coded by a domain expert or learned using bottom-up (Ciravegna, 2001; Califf & Mooney, 1999) or top-down learners (Soderland, 1999). Others follow statistical methods that define numerous features over the sentence and then treat the problem as a classification problem, applying well-known machine learning algorithms (e.g., Hidden Markov Models; Agichtein & Ganti, 2004; Borkar, Deshmukh, & Sarawagi, 2001). Our system does not deal with the many challenges in this field, as we propose a large scale domain-specific approach driven by specific extraction templates.

### 5.3.2 Relation Extraction

Relation extraction has been developed widely in the last years from large text corpora (Schubert, 2002) and, in particular, from different Web resources, such as general Web content (Banko et al., 2007; Carlson et al., 2010; Hoffmann, Zhang, & Weld, 2010), blogs (Jayram et al., 2006), Wikipedia (Suchanek et al., 2007), and news articles (e.g., the topic detection and tracking task (Section 5.3.3)). Given two entities, the first task in this domain is to classify their relationship. Many feature-based methods (Jiang & Zhai, 2007; Kambhatla, 2004; Suchanek, 2006) and rule-based methods (Aitken, 2002; Mcdonald, Chen, Su, & Marshall, 2004; Jayram et al., 2006; Shen et al., 2007) have been developed for this task. Most methods use different features extracted from the text, such as the words, the grammar features, such as parse tree and dependency graphs, and features extra ion from external relation repositories (e.g., Wikipedia Infobox) to add additional features (Nguyen & Moschitti, 2011; Hoffmann, Zhang, Ling, Zettlemoyer, & Weld, 2011). Labeled training examples, from which those feature are extracted, are then fed into a machine learning classifier, sometimes using transformations such as kernels (Zhao & Grishman, 2005; Zhang, Zhang, Su, & Zhou, 2006; Zelenko, Aone, & Richardella, 2003; Wang, 2008; Culotta & Sorensen, 2004; Bunescu & Mooney, 2005; Nguyen, Moschitti, & Riccardi, 2009), which, given new unseen entities, will be able to classify them into those categories.

Given a relation, the second common task in this domain is to find entities that satisfy this relation. Out of all information extraction tasks, this task is most relevant to ours, as we try to find structured events for which the causality relation holds. Most works in this domain focus on large collections, such as the Web, where labeling all entities and relations is infeasible (Agichtein & Gravano, 2000; Banko et al., 2007; Bunescu & Mooney, 2007; Rosenfeld & Feldman, 2007; Shinyama & Sekine, 2006; Turney, 2006). Usually seed entity databases are used, along with some manual extraction templates, and then expanded and filtered iteratively. Sarawagi states that "in spite of the extensive research on the topic, relationship extraction is by no means a solved problem. The accuracy values still range in the neighborhood of 50%–70% even in closed benchmark datasets ... In open domains like the Web, the state-of-the-art systems still involve a lot of special case handling that cannot easily be described as principled, portable approaches." (Sarawagi, 2008, p. 331). Similarly, in our task the size of our corpus does not allow us to assume any labeled sets. Instead, like the common approaches presented here, we also start with a predefined set of patterns.

### 5.3.3 TEMPORAL INFORMATION EXTRACTION

The temporal information extraction task deals with extraction and ordering of events from many events over time. Temporal information extraction can be categorized into three main subtasks – predicting the temporal order of events or time expressions described in text, predicting the relation between those events, and identifying when the document was written. This task has been found to be important in many natural language processing applications, such as question answering, information extraction, machine translation and text summarization, all of which require more than mere surface understanding. Most of these approaches (Ling & Weld, 2010; Mani, Schiffman, & Zhang, 2003; Lapata & Las-carides, 2006; Chambers et al., 2007; Tatu & Srikanth, 2008; Yoshikawa, Riedel, Asahara, & Matsumoto, 2009) learn classifiers that predict a temporal order of a pair of events from predefined features of the pair.

Other related works deal with topic detection and tracking (Cieri, Graff, Liberman, Martey, & Strassel, 2000). This area includes several tasks (Allan, 2002). In all of them, multiple, heterogenous new sources are used, including audio. The story segmentation task aims to segment data into its constituent stories. The topic tracking task – e.g., the work by Shahaf and Guestrin (2010) – aims to find all stories discussing a certain topic. A subtask of this is the link detection task which, given a pair of stories, aims to classify whether they are on the same topic. The topic detection task – e.g. the works by Ahmed, Ho, Eisenstein, Xing, Smola, and Teo (2011) and Yang, Pierce, and Carbonell (1998) – aims to detect clusters of topic-cohesive stories in a stream of topics. The first-story detection task aims to identify the first story on a topic (Jian Zhang & Yang, 2004). In this paper, we focused on short text headlines and the extraction of events from them. Our work differs from that of temporal information extraction, in that we generate predictions of future events, whereas temporal information extraction tasks focus on identifying and clustering the text corpus into topics.

### 5.3.4 CAUSALITY PATTERN EXTRACTION AND RECOGNITION

In the first stage of our learning process we extract causality pairs from text. Causality extraction has been discussed in the literature in the past, and can be divided into the following subgroups:

1. Use of handcrafted domain-specific patterns. Some studies deal with causality extraction using specific domain knowledge. Kaplan and Berry-Rogghe (1991) used scientific texts to create a manually designed set of propositions which were later applied on new texts to extract causality. These methods require handcrafted domain knowledge, which is problematic to obtain for real-world tasks, especially in large amounts.

2. Use of handcrafted linguistic patterns. These works use a more general approach by applying linguistic patterns. For example, Garcia (1997) manually identified 23 causative verb groups (e.g., to result in, to lead to, etc.). If a sentence contained one of those verbs, it was classified as containing a causation relation. A precision of 85% was reported. Khoo et al. (2000) used manually extracted graphical patterns based on syntactic parse trees, reporting accuracy of about 68% on an English medical database. Similarly, Girju and Moldovan (2002) defined lexicon-syntactic patterns (pairs of noun phrases with a causative verb in between) with additional semantic constraints.

3. Semi-supervised pattern learning approaches. This set of approaches uses supervised machine learning techniques to identify causality in text. For example, Blanco et al. (2008) and Sil et al. (2010) use syntactic patterns as features that are later fed into classifiers, whose output is whether the text implies causality or the cause and effect themselves.

4. Supervised pattern learning approaches. There have been many works on design inference rules to discover extraction patterns for a given relation using training examples (Riloff, 1996; Riloff & Jones, 1999; Agichtein & Gravano, 2000; Lin & Pantel, 2001). Specifically, Chan and Lam (2005) dealt with the problem of creating syntactic patterns for cause-effect extraction.

In the domain of causality pattern extraction, our work most resembles the handcrafted linguistic patterns pattern approaches. We evaluated their performance on our specific domain. Since our goal was to obtain a very precise set of examples to feed into our learning, we chose to follow such an approach as well.

## 5.4 Learning Causality

We have drawn some of our algorithmic motivation from work in the machine learning community. In this section, we give a partial review of the main areas of machine learning that are relevant to our work.

### 5.4.1 Bayesian Causal Inference

The functional causal model (Pearl, 2000) assumes a set of observables $X_1 \ldots X_n$, which are the vertices of a directed acyclic graph $G$. The semantics of the graph is that parents of a node are its directed causes. It was shown to satisfy Reichenbach's common cause principle, which states that for a node $Z$ with children $X$, $Y$, if $X$ and $Y$ are statistically dependent, then there is a $Z$ causally influencing both. This model, similar to a Bayesian network, satisfies several conditions: (1) Local Causal Markov condition: a node is statistically independent of non-descendants, given its parents; (2) Global Causal Markov condition: d-separation criterion; (3) Factorization criterion: $P(X_1, \ldots, X_n) = \prod_i P(X_i | Parents(X_i))$. The theoretical literature on the inference and learning of causality models is extensive. Those models resemble our work in the use of structural models. The literature on inference and learning of causality models is extensive, but to our knowledge there are no solutions that scale to the scope of tasks discussed in this paper. In contrast with Bayesian approach, the causality graph in our work contains less detailed information. Our work combines several linguistic resources that were learned from data with several heuristics to build the causality graph.

### 5.4.2 STRUCTURED LEARNING

An important problem in the machine learning field is structured learning, where the input or the output of the classifier is a complex structure, such as relational domain, where each object is related to another, either in time or in its features. Our task resembles structured learning in that we also use structured input (structured events given as input) and produce a structured event as output.

Many generative models have been developed, including hidden Markov models, Markov logic networks, and conditional random fields, among others. Other approaches use transformations, or kernels, that unite all the objects, ignoring the structure, and then feed it into a standard structured classifier, e.g., kernelized conditional random fields (Lafferty, Zhu, & Liu, 2004), maximum margin Markov networks (Taskar, Guestrin, & Koller, 2003), and others (Bakir, Hofmann, Schölkopf, Smola, Taskar, & Vishwanathan, 2007). When dealing with complex output, such as annotated parse trees for natural language problems, most approaches define a distance metric in the label space between the objects, and they again apply standard machine learning algorithms, e.g., structured support vector machines (Joachims, 2006).

### 5.4.3 LEARNING FROM POSITIVE EXAMPLES (ONE CLASS CLASSIFICATION)

As our system is only fed examples of the sort "a causes b," and no examples of the sort "a does not cause b," we must deal with the problem of learning from positive examples only. This is a challenge for most multi-class learning mechanisms, which require both negative and positive examples. Some theoretical studies of the possibility of learning from only positive unlabeled data are provided in the work by Denis (1998) (probably approximately correct (PAC) learning) and Muggleton (1996) (Bayesian learning).

Most works (Tax, 2001; Manevitz & Yousef, 2000; Manevitz, Yousef, Cristianini, Shawe-Taylor, & Williamson, 2001) in this domain develop algorithms that use one-class SVM (Vapnik, 1995) and learn the support using only positive distribution. They construct decision boundaries around the positive examples to differentiate them from all possible negative data. Tax and Duin (1991) use a hyper-sphere with some defined radius around some of the positive class points (support vector data description method). Some also use kernel tricks before finding this sphere (Tax, 2001). Schölkopf et al. (1999, 2000) develop methods that try to separate the surface region of the positive labeled data from the region of the unlabeled data.

## 6. Conclusions

Much research has been carried out on information extraction and ontology building. In this work, we discuss how to leverage such knowledge into a large-scale AI problem of event prediction. We present a system that is trained to predict future events, using a cause event as input. Each event is represented as a tuple of one predicate and 4 general semantic roles. The event pairs used for training are extracted automatically from news headlines using simple syntactic patterns. Generalization to unseen events is achieved by:

1. Creating an abstraction tree (AT) that contains entities from observed events together with their subsuming categories extracted from available online ontologies.

2. Finding predicate paths connecting entities from cause events to entities in the effect events, where the paths are again extracted from available ontologies.

We discuss the many challenges of building such a system: obtaining a large enough dataset, representing the knowledge, and developing the inference algorithms required for such a task. We perform *large-scale mining* and apply *natural language techniques* to transform the raw data of over 150 years of history archives into a structured representation of events, using a mined Web-based object hierarchy and action classes. This shows the scalability of the proposed method, which is crucial to any method that requires large amounts of data to work well. However, more engineering design and analysis should be performed to scale it to the entire knowledge of the web and provide real-time alerts. We also show that the numerous resources built by different people for different purposes (e.g., the different ontologies) can in fact be merged via a concept-graph to build a system that can work well in practice.

We perform *large-scale learning* over the large data corpus and present *novel inference* techniques. We consider both rule extraction and generalization. We propose novel methods for rule generalization using existing ontologies, which we believe can be useful for many other related tasks. Tasks such as entailment and topic tracking can benefit from the concepts of understanding sequences and their generalizations.

In this work we only scratch the surface of what can be a real-time fully functional prediction system. Due to the complexity of the problem, the size of the system and it many components, errors are unavoidable. For example, errors due to noise during event extraction, noise during the similarity calculation between events, etc. Although we perform experiments analyzing the different components of the system and their errors in addition to the overall system performance, we believe that additional training examples and better sources of knowledge and deeper ontologies can bring many improvements to our algorithms. For future work, we suggest the following directions and extensions:

1. Better event extraction and event matching – Event extraction techniques, e.g., as proposed by Do et al. (2011) can provide higher analysis of the data from the entire text rather than just the titles. Event similarity can be enriched in many ways, e.g., in this work we compared three aggregation functions $f$, however, a more coherent way of learning the weights of $O_i$ from past data can be applied.

2. Analysis of knowledge sources – We believe that more in-depth analysis of the different types of knowledge obtained from the Web and their individual contributions should be studied. In this work, we did not explore the sensitivity of the system to the initial noise of the conceptual networks, and we believe that proper analysis of those and better networks can provide higher prediction accuracy, as already being carried on by the LinkedData community.

3. Large scale experiments – Performance of larger experiments with humans over larger periods of times, and even comparison to experts can provide more insights on the performance and reliability of the system. Automation of such experiments without human involvement to measure accuracy of predictions will make it possible to provide richer metrics of performance, such as recall.

4. Time effect – In this work, all events were treated similarly, even events from 100 years ago. For future directions, we wish to investigate how to give decaying weight to information about events in the system, as causality learned from an event that took place in 1851 might be less relevant to the prediction in 2010. However, much common-sense knowledge can still be used even if learned from events that happened a long time ago. For example, the headlines "Order Restored After Riots" (1941) and "Games Suspended After Riot" (1962) are still relevant today.

Our experimental evaluation showed that the predictions of the Pundit algorithm are at least as good as those of non-expert humans. We believe that our work is one of the first to harness the vast amount of information available on the Web to perform event prediction that is general purpose, knowledge based, and human-like.